\documentclass[letterpaper, 10 pt, conference]{ieeeconf}  

\usepackage{hyperref}
\usepackage{amsmath,mathtools} 
\usepackage{amssymb}
\usepackage{graphicx} 
\usepackage{acronyms}
\usepackage{mfirstuc}
\usepackage{amsfonts}
\usepackage{xcolor} 
\usepackage{comment}
\usepackage{multirow}
\usepackage{algorithm}
\usepackage{siunitx}
\usepackage{algpseudocode}
\usepackage{afterpage}
\usepackage[font=small,labelfont=bf,figurename=Fig.]{caption} 
\usepackage{tikz}
\usetikzlibrary{positioning}
\usepackage{pgfplots}
\pgfplotsset{compat=1.18}
\usepackage{subcaption} 

\usepackage{float}
\usepackage{booktabs}
\usepackage{tabularray}
\usepackage[symbol]{footmisc} 
\usepackage[noadjust]{cite}

\IEEEoverridecommandlockouts                              

\title{\LARGE \bf
\capitalisewords{Uncertainty-Aware Adaptive Dynamics for Underwater Vehicle–Manipulator Robots}}

\author{Edward Morgan$^{1}$, Nenyi K Dadson$^{1}$,  Corina Barbalata$^{1, *}$
\thanks{*Corresponding author}
\thanks{$^{1}$The authors are with the Department of Mechanical \& Industrial Engineering,  at Louisiana State University, Baton Rouge, LA 70803, USA
        {\tt\small emorg31@lsu.edu, ndadso1@lsu.edu, cbarbalata@lsu.edu},}%
}

\begin{document}
\maketitle

\thispagestyle{empty}
\pagestyle{empty}

\begin{abstract}
Accurate and adaptive dynamic models are critical for underwater vehicle–manipulator systems where hydrodynamic effects induce time‐varying parameters. This paper introduces a novel uncertainty‐aware adaptive dynamics model framework that remains linear in lumped vehicle and manipulator parameters, and embeds convex physical consistency constraints during online estimation. Moving horizon estimation is used to stack horizon regressors, enforce realizable inertia, damping, friction, and hydrostatics, and quantify uncertainty from parameter evolution. Experiments on a BlueROV2 Heavy with a 4‐DOF manipulator demonstrate rapid convergence and calibrated predictions. Manipulator fits achieve $R^2=0.88$ to $0.98$ with slopes near unity, while vehicle surge, heave, and roll are reproduced with good fidelity under stronger coupling and noise. Median solver time is approximately $0.023$ s per update, confirming online feasibility. A comparison against a fixed parameter model shows consistent reductions in MAE and RMSE across degrees of freedom. Results indicate physically plausible parameters and confidence intervals with near 100\% coverage, enabling reliable feedforward control and simulation in underwater environments.
\end{abstract}

\section{Introduction}
\label{sec:introduction}

\acfp{UVMS} are versatile platforms for subsea operations such as inspection, intervention, and maintenance \cite{Morgan2022}. A six-degree-of-freedom vehicle combined with a multi-degree-of-freedom manipulator enables dexterous interaction, but operation is challenged by nonlinear hydrodynamics, strong vehicle–manipulator coupling, and time-varying uncertainties. These uncertainties arise because added mass, restoring forces, and other hydrodynamic effects alter the effective rigid-body parameters, leading to dynamics that vary with the surrounding fluid conditions \cite{10337120}. Accurate and adaptive models are therefore essential for control, planning, and reliable system monitoring.

Classical approaches often rely on fixed hydrodynamic coefficients or simplified coupling assumptions, which reduce complexity but fail to capture parameter variation induced by hydrodynamic influence. In contrast, regressor formulations offer a principled linear representation of dynamics in unknown parameters and provide a foundation for systematic identification and adaptive control. However, their application in \ac{UVMS} has been limited by the challenge of updating parameters online while simultaneously guaranteeing physical plausibility.

\begin{figure}[!t]
    \centering
    \includegraphics[width=0.5\textwidth]{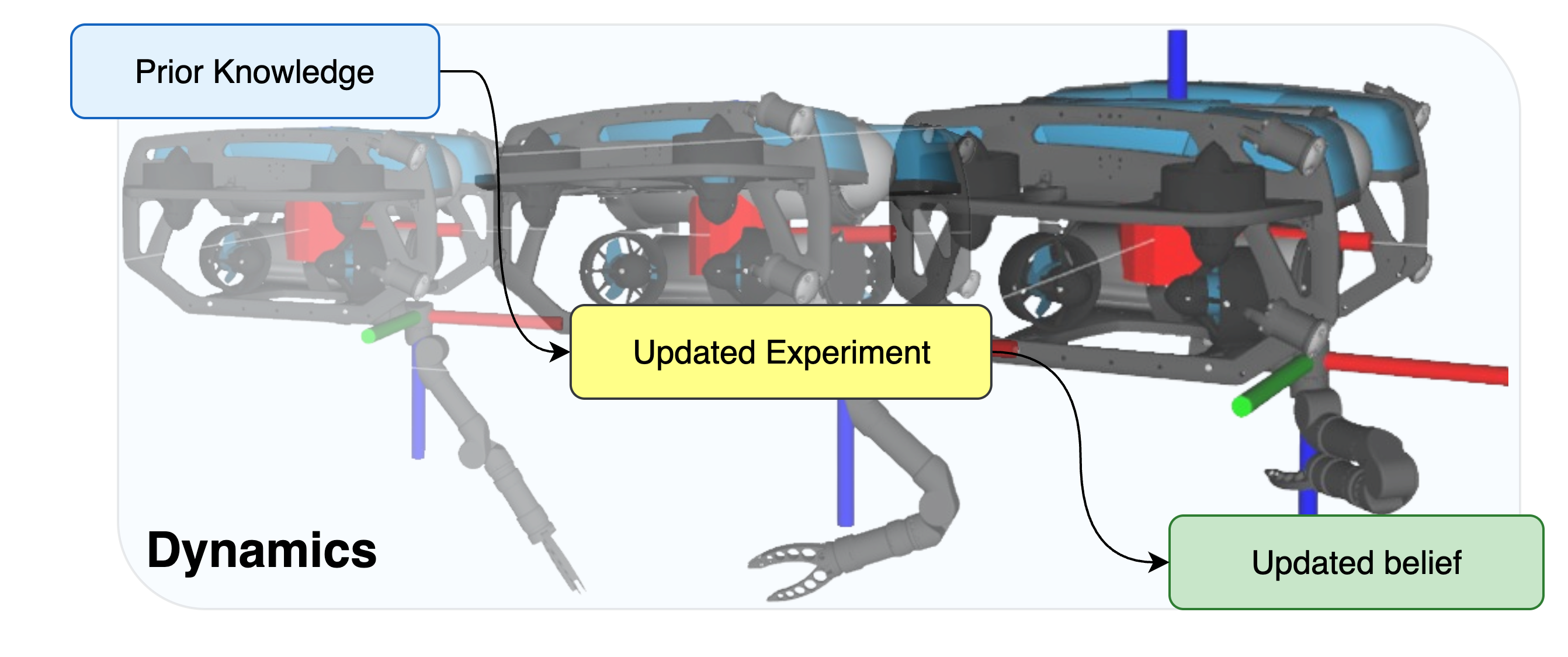}
  \caption{Adaptive dynamics in UVMS: prior knowledge of system parameters is updated using new experimental data, resulting in an updated belief consistent with observed dynamics.}
  \label{fig:idea}
\end{figure}

Recent advances in convex optimization enable physical consistency constraints to be embedded directly in online estimation. This capability makes it possible to design models that adapt parameters under hydrodynamic variation while ensuring realizable inertia, damping, friction, and hydrostatics. Motivated by this, an adaptive regressor-based framework for UVMS dynamics, \figurename~\ref{fig:idea}, is proposed, combining moving horizon estimation with convex physical constraints and uncertainty quantification from parameter changes.

\begin{figure*}[ht]
  \centering
  \includegraphics[width=\textwidth]{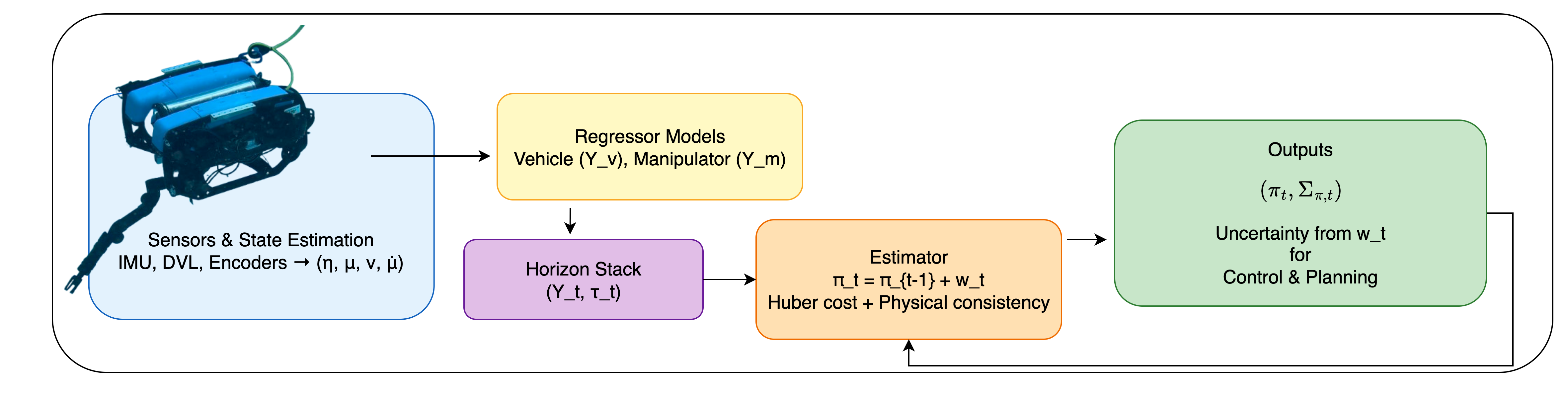}
  \caption{Framework overview of the Uncertainty-Aware Adaptive Dynamics scheme for UVMS, combining regressor models, horizon stacking, moving horizon estimation, and physical consistency constraints to produce adaptive parameter estimates with uncertainty for control and planning.}
  \label{fig:flow}
\end{figure*}

\subsection{Related Work}
Classical marine dynamics provide the standard 6-DOF vehicle model, which captures added mass, Coriolis and centripetal effects, hydrodynamic damping, and hydrostatic restoring forces \cite{fossen2011}. This model serves as the foundation for the vehicle dynamics considered here. For \ac{UVMS} modeling and control, established references extend this foundation to describe kinematic and dynamic coordination between vehicle and manipulator, as well as interaction control and task prioritization \cite{antonelli2014,sivcev2018}.

In robotics, regressor-based formulations exploit the linear dependence of dynamics on unknown inertial and dissipative parameters, enabling identification and feedforward control \cite{slotine1987,atkeson1986,khosla1985}. To guarantee physical plausibility, convex optimization methods have been introduced that enforce positive-definite inertias and feasible mass distributions through pseudo-inertia matrices and linear matrix inequalities \cite{traversaro2016,wensing2018}.

For underwater vehicles specifically, parameter identification has been studied in both offline and online settings, often using excitation strategies with DVL, IMU, and depth sensors \cite{smallwood2003,ridao2004,kinsey2006,avila2008}. However, most studies address the vehicle in isolation, without leveraging block-structured regressors for coupled vehicle–manipulator dynamics or enforcing physical consistency constraints during online adaptation.

Moving horizon estimation provides a natural framework for this problem, as it supports constrained optimization-based updates over finite horizons and admits robust penalties to reduce the effect of outliers \cite{rao2001,rawlings2017,huber1964}. Yet, a unified approach that stacks vehicle and manipulator regressors over a horizon, enforces convex physical consistency, and propagates calibrated uncertainty from parameter increments remains underexplored. This gap motivates the Uncertainty-Aware Adaptive Dynamics framework presented in this work, shown in \figurename~\ref{fig:flow}.

\subsection{Contribution}

This work presents an \emph{uncertainty-aware adaptive regressor-based modeling framework} for underwater vehicle–manipulator systems. The main contributions are:
\begin{itemize}
    \item A unified regressor formulation for coupled UVMS dynamics that remains linear in lumped inertial, restoring, damping, and friction parameters across vehicle and manipulator subsystems.  
    \item An online moving horizon estimation scheme that adapts parameters online under hydrodynamic variation while enforcing convex constraints on inertia, friction, and hydrostatics for physical consistency.  
    \item An uncertainty quantification mechanism based on observed parameter changes, providing interpretable confidence intervals for adaptive models.  
    \item Experimental validation that the proposed approach improves model fidelity, highlighting its utility for prediction and adaptive control in underwater environments.  
\end{itemize}

The remainder of this paper is organized as follows. Section~\ref{sec:prob_statement} presents the problem formulation and the UVMS dynamic model. Section~\ref{sec:methodology} details the regressor-based modeling and adaptive estimation methodology. Section~\ref{sec:results} reports experimental validation and analysis. Section~\ref{sec:conclusions} concludes with key findings and directions for future research.

\section{Problem Formulation}
\label{sec:prob_statement}
\subsection{UVMS Description}
The dynamics of a \ac{UVMS} is expressed as
\begin{equation}
M(\zeta, \pi)\,\ddot{\zeta}
+ h(\zeta, \dot{\zeta}, \pi)
= \tau,
\label{eq:system_dynamics}
\end{equation}
where $M \in \mathbb{R}^{(6+n)\times(6+n)}$ is the generalized inertia matrix, 
$h \in \mathbb{R}^{(6+n)\times1}$ collects Coriolis, damping, restoring, and friction effects,  $\tau \in \mathbb{R}^{(6+n)\times1}$ stacks a $6$-\ac{DOF} vehicle forces and moments and $n$-\ac{DOF} manipulator torques, and $\pi$
represents the uncertain rigid body and hydrodynamic parameters. 

The generalized state vector is defined with
$
{\zeta =
\begin{bmatrix}
\eta,
\mu
\end{bmatrix}^T}, \
\dot{\zeta} =
\begin{bmatrix}
\nu, 
\dot{\mu}
\end{bmatrix}^T,
$
where $\eta\in\mathbb{R}^6$ is the vehicle pose, $\mu\in\mathbb{R}^n$ are the manipulator joint angles, and $\nu=[u,v,w,p,q,r]^T$ is the body-fixed velocity of the vehicle. The kinematic relation
$
\dot{\eta} = J(\eta)\,\nu
\label{eq:zeta_dot}
$
maps vehicle body-fixed velocity to Euler-angle rates.

The uncertain parameters are grouped as
$
{\pi =
\begin{bmatrix}
\pi_v, 
\pi_m
\end{bmatrix}^T}, 
\ 
\pi_v \in \mathbb{R}^{k},\;
\pi_m \in \mathbb{R}^{12n}
$, where
 $k$ is the number of lumped parameters
for the 6-DOF vehicle model. A full description of $\pi_v$ would require many
hydrodynamic derivatives, which is impractical, but body-symmetry
considerations allow the number of unknowns to be reduced substantially.

$\pi_v$ collects lumped vehicle contributions:  
$(i)$ effective inertial parameters (rigid-body mass and inertia with added-mass effects),  
$(ii)$ linear and quadratic drag coefficients, and  
$(iii)$ restoring-force parameters (weight, buoyancy, and center-of-gravity/center-of-buoyancy offsets).  
The manipulator parameters $\pi_m$ include  
$(i)$ effective link inertial parameters (rigid-body and added-mass terms), and  
$(ii)$ joint-level dissipative terms (viscous friction, Coulomb friction, and hydrodynamic drag mapped into joint space).  

This lumped parameterization preserves linearity in the dynamics for identification and control, while full physical quantities can be reconstructed from $\pi$ together with known constants such as vehicle and manipulator weight.

\subsection{Vehicle Subsystem}
The vehicle subsystem is described by
\begin{equation}
M_v(\pi_v)\,\dot{\nu}
+ C_v(\nu, \pi_v)\,\nu
+ D_v(\nu, \pi_v)\,\nu
+ g_v(\eta, \pi_v)
= \tau_v + \tau_{mv},
\label{eq:vehicle_dynamics_uncertain}
\end{equation}
where $M_v \in \mathbb{R}^{6\times6}$ is the  inertia matrix including rigid-body and added-mass terms, $C_v \in \mathbb{R}^{6\times6}$ is the Coriolis–centripetal matrix, $D_v \in \mathbb{R}^{6\times6}$ embeds linear and quadratic drag in body coordinates, $g_v \in \mathbb{R}^{6\times1}$ accounts for restoring forces, $\tau_v \in \mathbb{R}^{6\times1}$ is the forces and moments for control, and $\tau_{mv} \in \mathbb{R}^{6\times1}$ is the manipulator reaction wrench.

\subsection{Manipulator Subsystem}
For an $n$-DOF manipulator, the joint-space dynamics are
\begin{equation}
M_m(\mu, \pi_m)\,\ddot{\mu}
+ C_m(\mu, \dot{\mu}, \pi_m)\,\dot{\mu}
+ f_m(\dot{\mu}, \pi_m)
+ g_m(\mu, \pi_m)
= \tau_m,
\label{eq:manipulator_dynamics_uncertain}
\end{equation}
where $M_m \in \mathbb{R}^{n\times n}$ is the effective link inertia matrix including rigid-body and added-mass contributions, $C_m \in \mathbb{R}^{n\times n}$ captures centrifugal and Coriolis effects, $f_m(\dot{\mu}, \pi_m) \in \mathbb{R}^{n\times 1}$ represents dissipative effects at the joint level, including viscous friction, Coulomb friction, and hydrodynamic drag transformed from link to joint space, $g_m \in \mathbb{R}^{n\times 1}$ accounts for gravity and buoyancy, and $\tau_m \in \mathbb{R}^{n\times 1}$ are the applied joint torques.

\subsection{Problem Statement}
The objective of this work is to estimate the time-varying parameter vector $\pi$ of the coupled \ac{UVMS} online, using measured states and inputs over a finite horizon, while also providing a calibrated measure of parameter uncertainty. The estimator returns $(\pi_t, \Sigma_{\pi,t})$, where $\pi_t$ denotes the adapted parameters and $\Sigma_{\pi,t}$ quantifies their uncertainty. The constraint of this objective are:
\begin{itemize}
    \item Preserve linearity in the lumped parameterization to enable efficient regressor-based identification,
    \item Adapt to variations in inertia, damping, and restoring forces induced by hydrodynamic effects,
    \item Enforce physical consistency constraints on inertia, dissipative effects, and hydrostatics to guarantee realizable dynamic models.
\end{itemize}

\section{Methodology}
\label{sec:methodology}

\subsection{Underwater Vehicle Regressor Formulation}

The vehicle dynamics \eqref{eq:vehicle_dynamics_uncertain} can be expressed in regressor form
\begin{equation}
\tau_v + \tau_{mv} = Y_v(\nu, \dot{\nu}, \eta)\,\pi_v
\label{eq:vehicle_dynamics_uncertain_linear}
\end{equation}

\subsubsection*{Lumped Parameterization}
The vehicle parameter vector is partitioned into three groups,  
$\pi_v = [\,\pi_{vM};\, \pi_{vD};\, \pi_{vG}\,]^T$, where  
$\pi_{vM} = [\,m - X_{du}, \ldots, I_z - N_{dr}\,] \in \mathbb{R}^{k_m}$ collects
effective inertial terms (rigid-body mass and inertia combined with added-mass effects),  
$\pi_{vD} = [\,X_u, \ldots, N_{rr}\,] \in \mathbb{R}^{k_d}$ contains linear and quadratic
damping coefficients in body coordinates, and  
$\pi_{vG} = [\,W, B, \ldots, z_gW - z_bB\,] \in \mathbb{R}^{k_g}$ includes restoring
force parameters such as weight $W$, buoyancy $B$, and center-of-gravity/center-of-buoyancy
offsets.  
Thus, the vehicle lumped parameter dimension is $k = k_m + k_d + k_g$.

\subsubsection*{Underwater Vehicle Regressor}
The vehicle regressor is
\[
Y_v = [\,Y_{vM};\; Y_{vD};\; Y_{vG}\,],
\]
where $Y_{vM} \in \mathbb{R}^{6\times k_m}$ is the Jacobian of inertial and Coriolis
terms $M_v\dot{\nu}+C_v(\nu)\nu$ w.r.t.~$\pi_{vM}$,  
$Y_{vD} \in \mathbb{R}^{6\times k_d}$ is the Jacobian of the drag forces $D_v(\nu,\pi_v)$
w.r.t.~$\pi_{vD}$, and  
$Y_{vG} \in \mathbb{R}^{6\times k_g}$ is the Jacobian of the restoring forces
$g_v(\eta,\pi_v)$ w.r.t.~$\pi_{vG}$. Thus, the complete vehicle regressor has dimension $Y_v \in \mathbb{R}^{6\times k}$.  
This preserves linearity in the unknown parameters while accommodating nonlinear
dependence on the vehicle states.

\subsection{Manipulator Regressor Formulation}

Manipulator dynamics \eqref{eq:manipulator_dynamics_uncertain} are expressed in regressor form \cite{10.5555/1524151}
\[
\tau_m = Y_m(\mu, \dot{\mu}, \ddot{\mu}, g)\,\pi_m,
\]
where $\pi_m = [\,\pi_1;\ldots;\pi_n\,]^T \in \mathbb{R}^{12n}$ stacks the link-level parameters.

\subsubsection*{Link-Level Parameters}
Each link $i$ contributes $12$ effective parameters: the mass $m_i$; the first
moments $m_i\ell_{x_i}, m_i\ell_{y_i}, m_i\ell_{z_i}$ with $\ell_{x_i},\ell_{y_i},\ell_{z_i}$
the \ac{COM} coordinates in the link frame; the inertia components
$I_{xx,i}, I_{yy,i}, I_{zz,i}, I_{xy,i}, I_{xz,i}, I_{yz,i}$ about the link
frame origin; and two joint-level dissipative coefficients $f_{v,i}, f_{s,i}$ that lump
friction and hydrodynamic drag mapped into joint space. All quantities are
\emph{effective parameters}, combining rigid-body properties with hydrodynamic
added-mass and drag effects.

\subsubsection*{Regressor Structure}
The torque at joint $i$ follows $\tau_{m,i} = \sum_{j=1}^n y_{ij}^T\pi_j$, with
$
y_{ij} =
\begin{bmatrix}
\tilde{y}_{ij} \\[4pt]
\delta_{ij}\,\phi_i(\dot{\mu}_i)
\end{bmatrix}
\in \mathbb{R}^{12},
$
where
$
\tilde{y}_{ij} = \tfrac{d}{dt}\!\left(\tfrac{\partial \beta_{T_j}}{\partial \dot{\mu}_i}\right)
- \tfrac{\partial \beta_{T_j}}{\partial \mu_i}
+ \tfrac{\partial \beta_{U_j}}{\partial \mu_i},
$
with $\beta_{T_j},\beta_{U_j}$ the kinetic and potential energy regressors of link $j$, and 
$\phi_i(\dot{\mu}_i) = [\,\dot{\mu}_i,\; \operatorname{sgn}(\dot{\mu}_i)\,]^T$, 
$\delta_{ij}=1$ if $i=j$ and $0$ otherwise, so the dissipative terms remain local to each joint.

Since each $\beta_{T_j},\beta_{U_j}$ depends only on joints $1,\ldots,j$, it follows that $y_{ij}=0$ for $j<i$, yielding the block upper-triangular regressor
\[
Y_m =
\begin{bmatrix}
y_{11}^T & y_{12}^T & \cdots & y_{1n}^T\\
0^T & y_{22}^T & \cdots & y_{2n}^T\\
\vdots & \vdots & \ddots & \vdots\\
0^T & 0^T & \cdots & y_{nn}^T
\end{bmatrix} \in \mathbb{R}^{n\times 12n}.
\]

\subsection{Complete System Regressor}

Combining the vehicle and manipulator regressors yields the system-level formulation consistent with \eqref{eq:system_dynamics}:
\begin{equation}
\begin{split}
\tau &= Y(\zeta, \dot{\zeta}, \ddot{\zeta})\pi, \\ 
\begin{bmatrix}
\tau_v + \tau_{mv} \\[4pt]
\tau_m
\end{bmatrix}
& =
\begin{bmatrix}
Y_v & 0 \\
0 & Y_m
\end{bmatrix}
\begin{bmatrix}
\pi_v \\[3pt]
\pi_m
\end{bmatrix}
\label{eq:system_regressor}
\end{split}
\end{equation}

Equation \eqref{eq:system_regressor} defines the complete UVMS regressor, with $Y_v$ capturing vehicle inertial, damping, and restoring contributions, $Y_m$ capturing manipulator link dynamics, and the coupling forces $\tau_{mv}$ naturally included in the vehicle part. This block-structured formulation preserves linearity in the parameter vector $\pi$ and highlights how the subsystem regressors integrate into the full coupled model, providing the foundation for the adaptive dynamics formulation.

\subsection{Parameter Estimation}
A moving horizon estimation (MHE) scheme is employed to adaptively estimate $\pi$ online. This work uses a robust convex formulation to handle non-Gaussian disturbances and outliers, while enforcing physically plausible parameter estimates.

The parameter estimation is cast as a discrete-time linear dynamical system with state $\pi_t \in \mathbb{R}^{k + 12n}$ at time $t$:
\begin{equation}
\begin{aligned}
\pi_{t} &= \pi_{t-1} + w_t, \\
\tau_{t} &= Y_t \pi_{t} + v_t,
\end{aligned}
\label{eq:adapt_dynamics}
\end{equation}
where $w_t \in \mathbb{R}^{(k+12n) \times 1}$ represents the change in parameters due to adaptation, $v_t \in \mathbb{R}^{N(6+n) \times 1}$ is the $N$ horizon-stacked modeling residual, and $Y_t, \tau_t$ denote the horizon-stacked regressors and observed forces/torques.

\subsubsection*{Optimization Problem}
Given stacked observations over a finite horizon, the estimate $\pi_t$ is obtained by solving the convex problem:
\begin{equation}
\begin{array}{ll}
\text{minimize} & \|w_t\|_Q^2 +  \phi_\rho(v_t) \\[6pt]
\text{subject to} &
\begin{cases}
\pi_t = \pi_{t-1} + w_t, \\
\tau_t = Y_t \pi_t + v_t, \\
\pi_t \in \mathcal{P},
\end{cases}
\end{array}
\label{eq:adapt_optimization}
\end{equation}
where $\|w_t\|_Q^2 = w_t^T Q w_t$ with $Q \succ 0$ a scaling matrix, and $\phi_\rho(\cdot)$ is the Huber penalty
\begin{equation}
\phi_\rho(v_t) =
\begin{cases}
\|v_t\|_2^2, & \|v_t\|_2 \leq \rho, \\[4pt]
2\rho\|v_t\|_2 - \rho^2, & \|v_t\|_2 > \rho,
\end{cases}
\end{equation}
which limits the impact of large outliers.


\subsubsection*{Physical Consistency Constraints}
The feasible set $\mathcal{P}$ encodes physically plausible conditions on the estimated parameters:
\begin{equation}
\mathcal{P} = \left\{ \pi \,\middle|\, 
\begin{aligned}
& M_v(\pi_v) \succ 0,\; M_v(\pi_v)^T = M_v(\pi_v) \\
& J_j(\pi_{j}) \succ 0,\quad j=1,\ldots,n \\
& f_{v,i} \ge 0,\; f_{s,i} \ge 0,\quad i=1,\ldots,n \\
& \pi_{vD,\mathrm{lin}} \le 0,\;\; \pi_{vD,\mathrm{quad}} \le 0 \\
& W_{\max} \ge W \ge W_{\min}
\end{aligned}
\right\}
\end{equation}
where $J_j(\pi_{j})$ and $M_v(\pi_v)$ denote the manipulator pseudo inertia and vehicle inertia matrices. The coefficients $f_{v,i}$ and $f_{s,i}$ are viscous and Coulomb dissipation coefficients, $\pi_{vD,\mathrm{lin}}$ and $\pi_{vD,\mathrm{quad}}$ are vehicle linear and quadratic damping terms, and $W$ is the vehicle weight. The formulation is convex with affine equalities and convex inequalities since parameters enter linearly.

\subsection{Uncertainty Quantification}
Uncertainty in the parameter estimates is quantified from the sequence of increments $w_t$ in the adaptive law ${\pi_t = \pi_{t-1} + w_t}$.
An \emph{exponentially weighted covariance} is employed to capture both variability and drift in the parameters. Since the parameter set consist of heterogeneous quantities of different magnitudes, these are normalized to a common scale. Defining $s_{t-1} = \max(|\pi_{t-1}|,\,\varepsilon)$ as the maximum parameter value, 
with $\varepsilon>0$ to avoid division by zero, the normalization of the increments  is given by  $
\tilde w_t = s_{t-1}^{-1} \odot w_t$. The mean $\bar{\tilde w}_t$ and covariance $\Sigma^{(w)}_t$ are updated as:

\begin{align}
\bar{\tilde w}_t &= (1-\alpha)\,\bar{\tilde w}_{t-1} + \alpha\,\tilde w_t, \\[2pt]
\Sigma^{(\tilde w)}_t &= (1-\alpha)\,\Sigma^{(\tilde w)}_{t-1} + \alpha\,(\tilde w_t-\bar{\tilde w}_{t-1})(\tilde w_t-\bar{\tilde w}_t)^{T} + \epsilon I,
\end{align}
with an importance factor $\alpha \in (0,1]$.  

Mapping back to the actual scale of the parameters yields the increment covariance
$
\Sigma^{(w)}_t = S_{t-1}$,$\Sigma^{(\tilde w)}_t\,S_{t-1}, $ ${S_{t-1}=\operatorname{diag}(s_{t-1})},
$
which reflects the variance of the updates. To compute the covariance $\Sigma_{\pi,t}$ of the parameters themselves, $\pi_t$, scaling is done by $L \approx 2/\alpha - 1$, which accounts for the contribution of the importance factor, $\alpha$. This results in:
$
\Sigma_{\pi,t} \;\approx\; L\,\Sigma^{(w)}_t.
$

\subsection{Use Cases of Estimated Parameters}
\label{sec:param_propagation}

The estimated parameters $\pi_t$ are used in both inverse and forward dynamics to demonstrate the importance of such approaches for creating accurate simulation models or feed-forward control methods.

\subsubsection*{Inverse Dynamics}
Given states $(\zeta_t,\dot{\zeta}_t,\ddot{\zeta}_t)$, forces and torques are computed as
$
\tau_t = Y_t\,\pi_t,
$
linear in $\pi_t$ and useful for feedforward control.  
Uncertainty from the parameters propagates as
$\Sigma_{\tau,t}^{\mathrm{fit}} \approx Y_t\,\Sigma_{\pi,t}\,Y_t^T$. In practice, predictive variance must also account for unmodeled effects and sensor noise. If $\Sigma_{\tau,t}^{\text{noise}}$ denotes a diagonal covariance estimated from residual statistics, then
$
\Sigma_{\tau,t}^{\text{pred}} \;\approx\; \Sigma_{\tau,t}^{\text{fit}} + \Sigma_{\tau,t}^{\text{noise}}
$
provides confidence bands relative to actual torque measurements.

\subsubsection*{Forward Dynamics} Given commanded torques $\tau_c$ and initial states, the proposed approach can be used for creating accurate digital-twins of dynamic models. In this case $\pi_t$ reconstructs the physical matrices $M,h$, yielding
\begin{equation}
\begin{aligned}
M(\zeta,\pi_t)\ddot{\zeta} = \tau - h(\zeta,\dot{\zeta},\pi_t) \\
\ddot{\zeta} = M(\zeta,\pi_t)^{-1}(\tau - h(\cdot)).
\end{aligned}
\end{equation}
Acceleration variance can be approximated by linearizing $a = \ddot{\zeta}(\cdot)$ so that
$
\Sigma_{\ddot{\zeta},t} \approx
\Big(\tfrac{\partial a}{\partial \pi}\Big)\,\Sigma_{\pi,t}\,\Big(\tfrac{\partial a}{\partial \pi}\Big)^{T}.
$
Including residual disturbance variance yields predictive uncertainty on forward simulated states.

\vspace{-5pt}
\section{Experiment \& Results}
\label{sec:results}
\vspace{-2pt}

\begin{figure*}[htb]
  \centering
  \begin{subfigure}{0.32\textwidth}
    \centering
    \includegraphics[width=\linewidth]{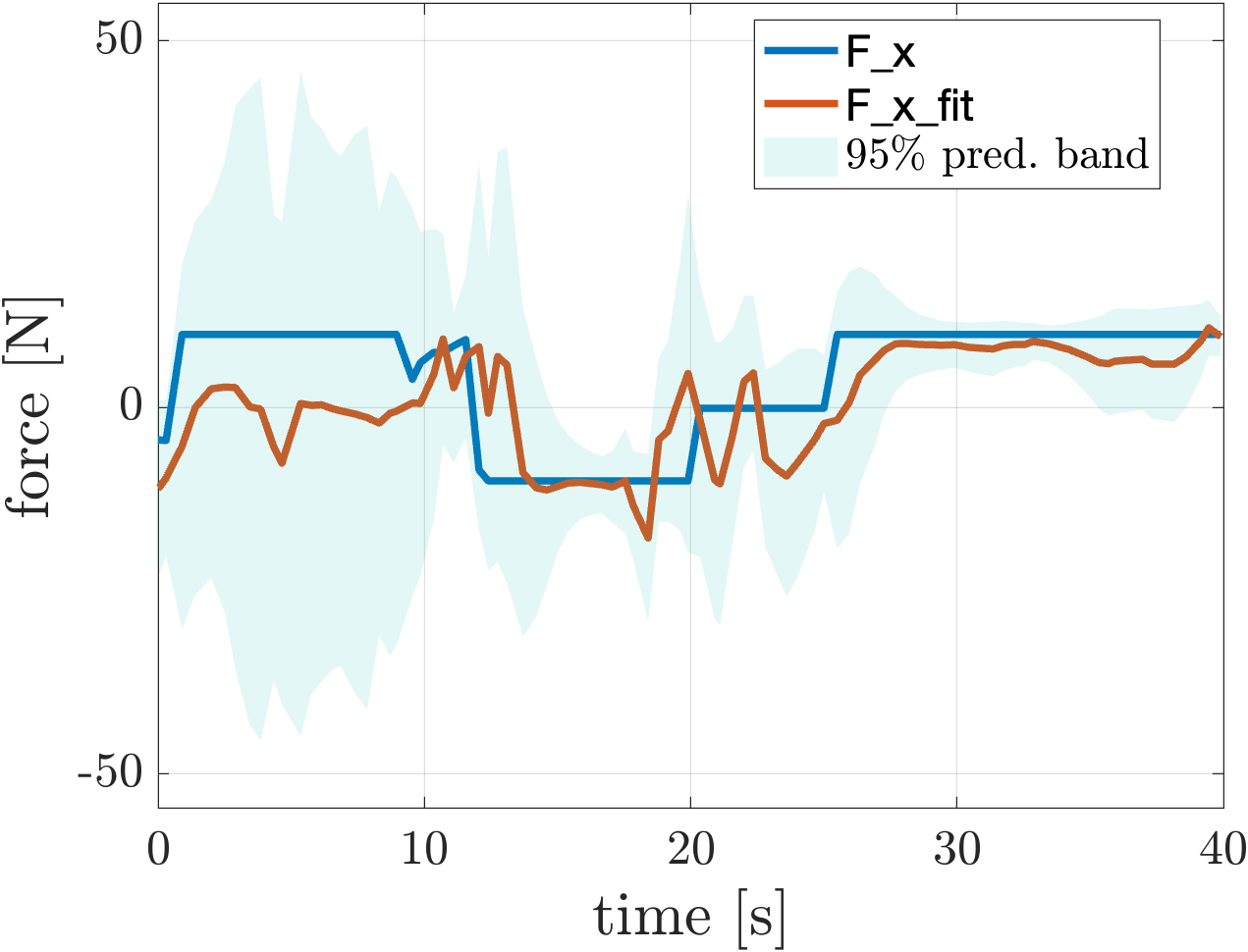}
    \label{fig:fit_vehicle_x}
  \end{subfigure}
  ~
  \begin{subfigure}{0.32\textwidth}
    \centering
    \includegraphics[width=\linewidth]{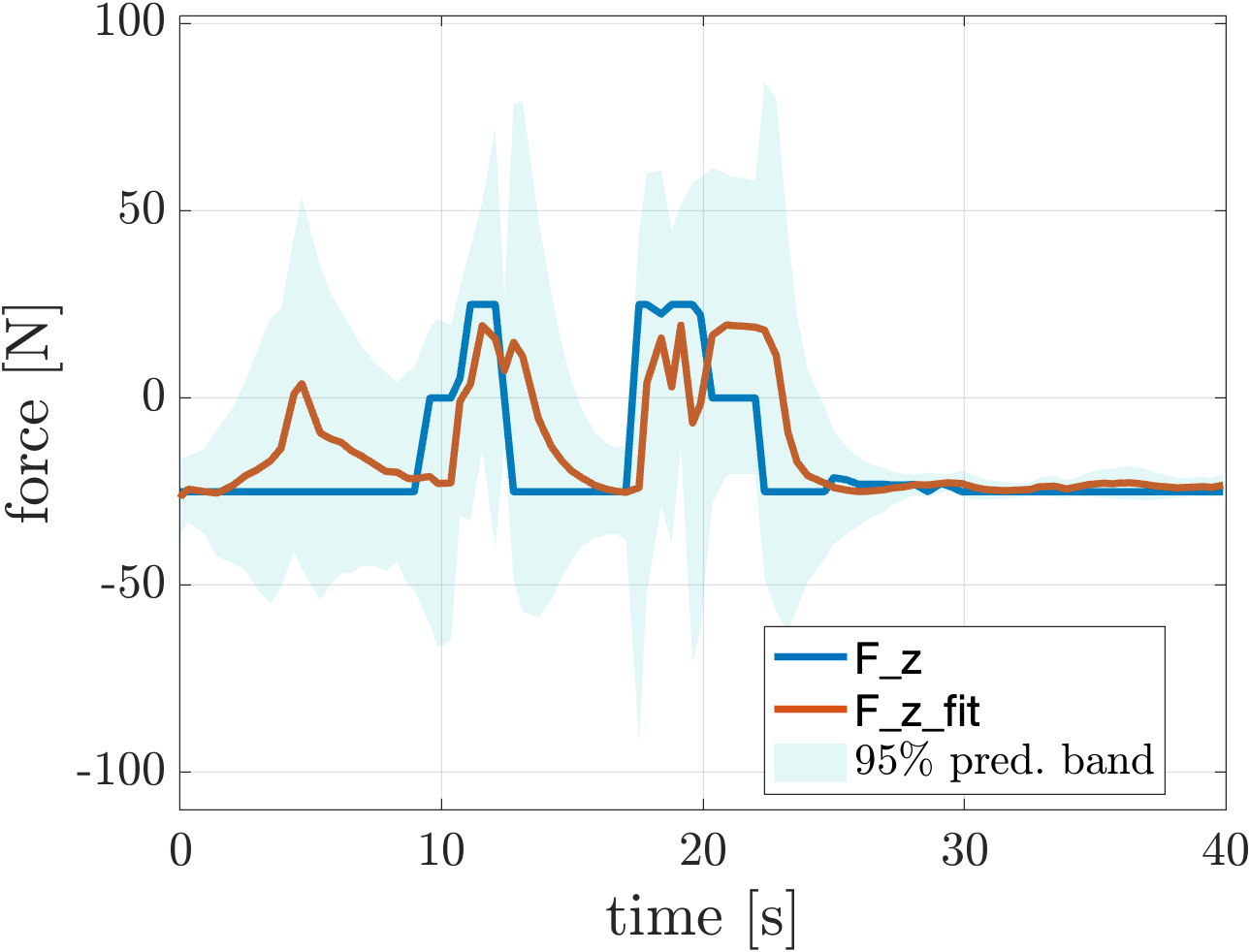}
    \label{fig:fit_vehicle_z}
  \end{subfigure}
    ~
  \begin{subfigure}{0.32\textwidth}
    \centering
    \includegraphics[width=\linewidth]{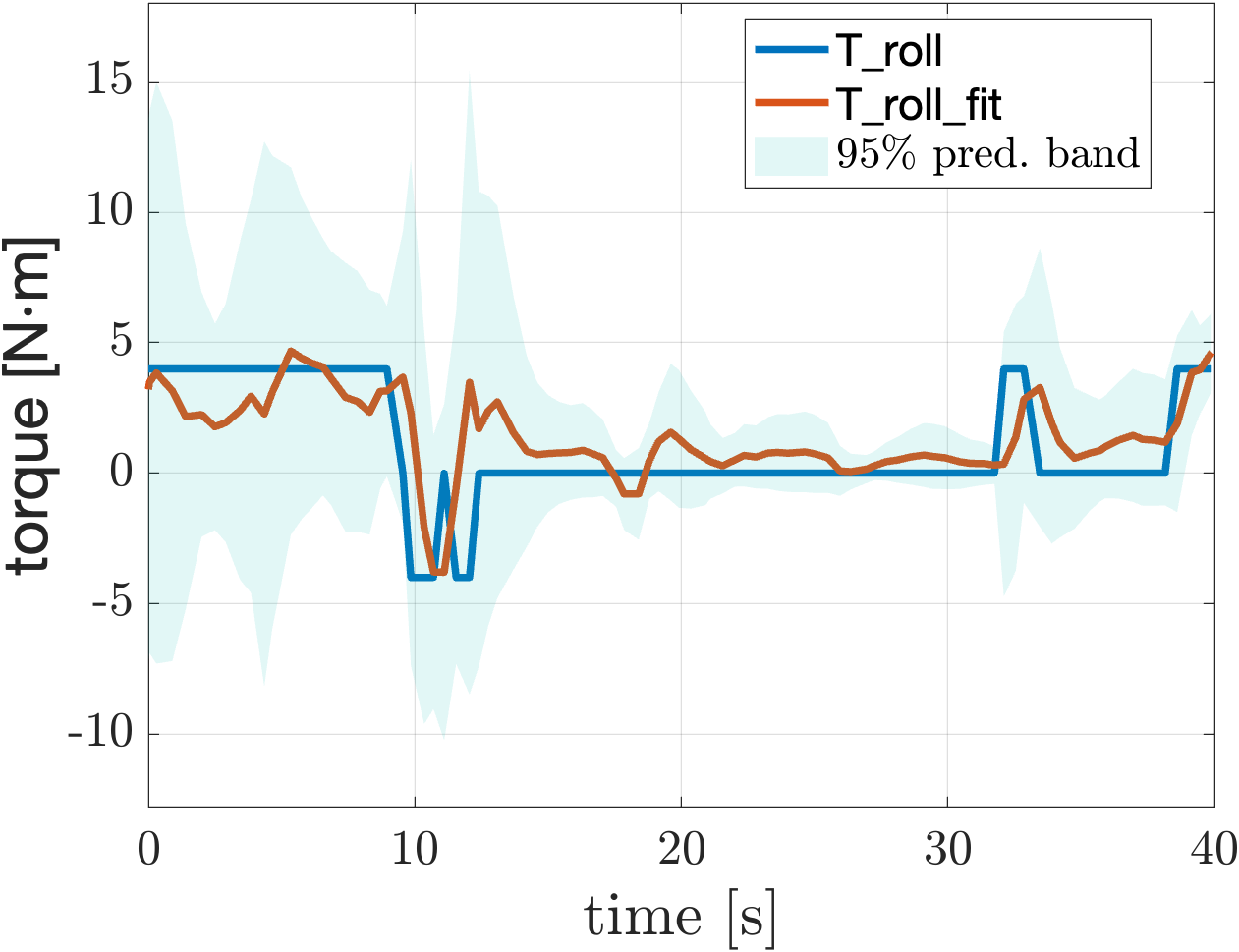}
  \end{subfigure}
  \caption{Identified fit for excited and recorded vehicle  (a) surge force, (b) heave force and (c) roll moment \acs{DOF}.}
  \label{fig:vehicle-prediction}
\end{figure*}

\begin{figure*}[htb]
  \centering
  \begin{subfigure}{0.4\textwidth}
    \centering  \includegraphics[width=\linewidth]{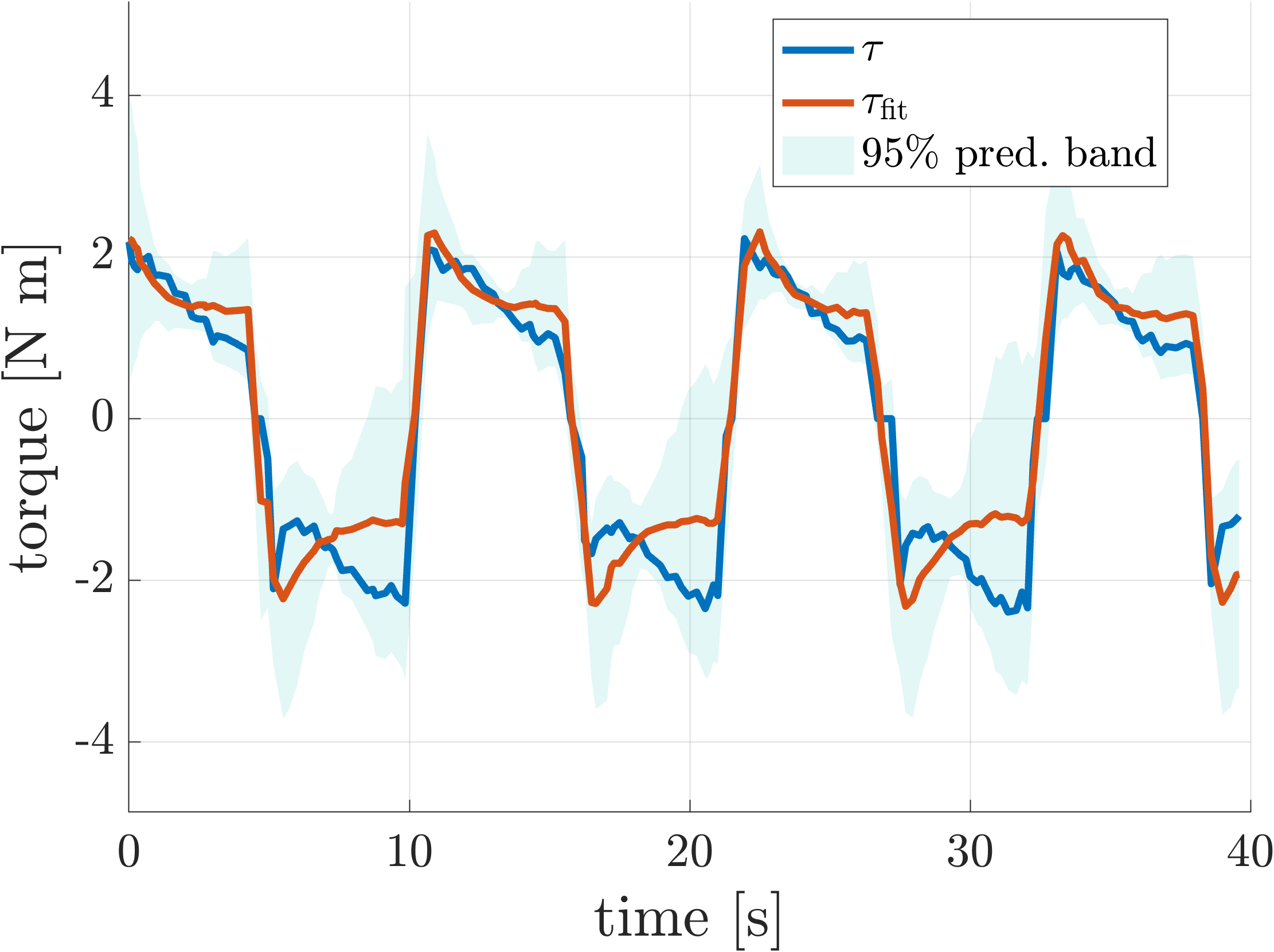}
  \end{subfigure}
  \begin{subfigure}{0.4\textwidth}
    \centering \includegraphics[width=\linewidth]{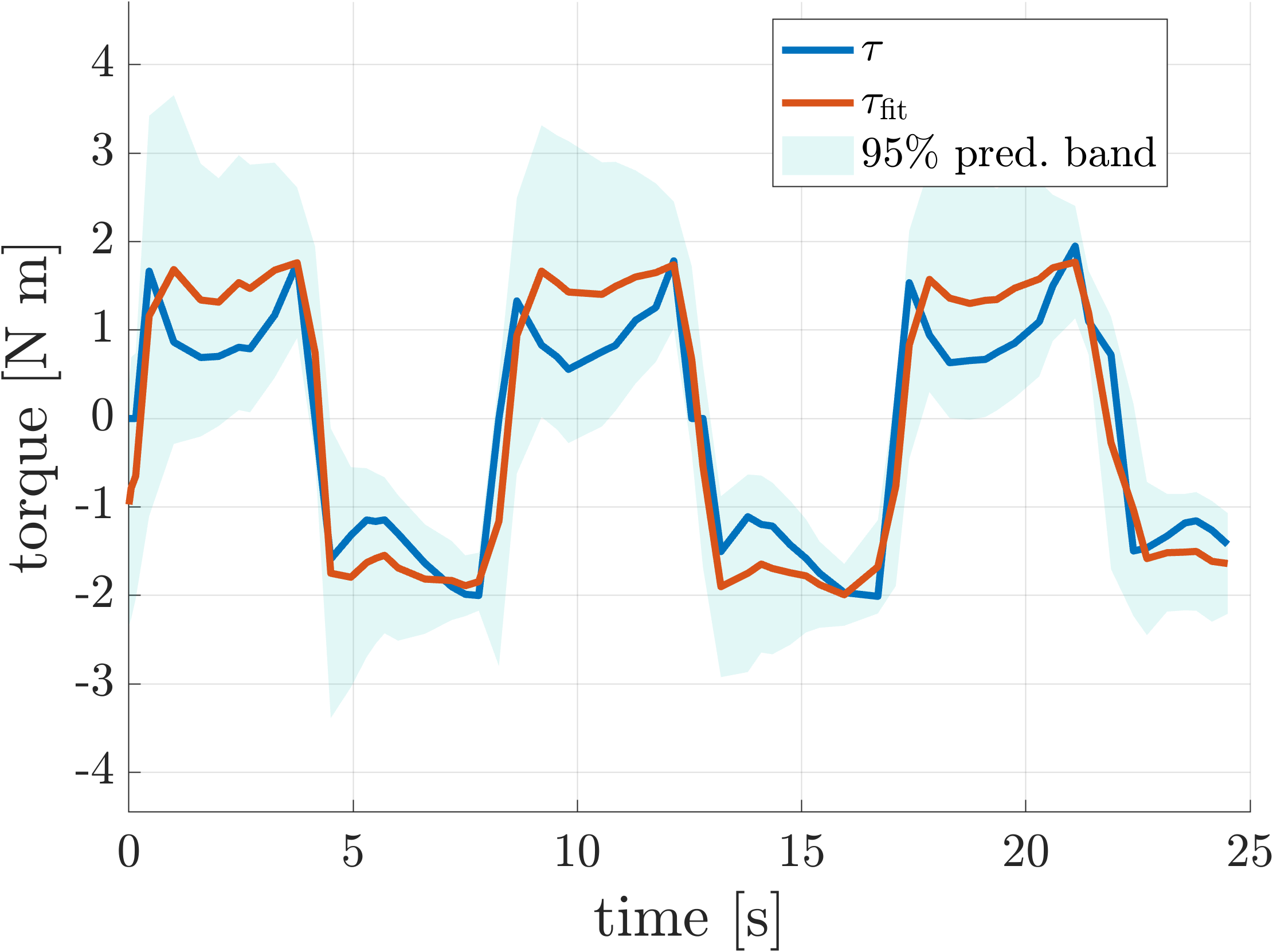}
  \end{subfigure}
  \caption{Identified fit for excited and recorded torque profile of manipulator: (a) joint 0  and (b) joint 1.}
  \label{fig:fit_joint_0}
\end{figure*}

\begin{figure*}[htb]
  \centering
  \begin{subfigure}{0.4\textwidth}
    \centering    \includegraphics[width=\linewidth]{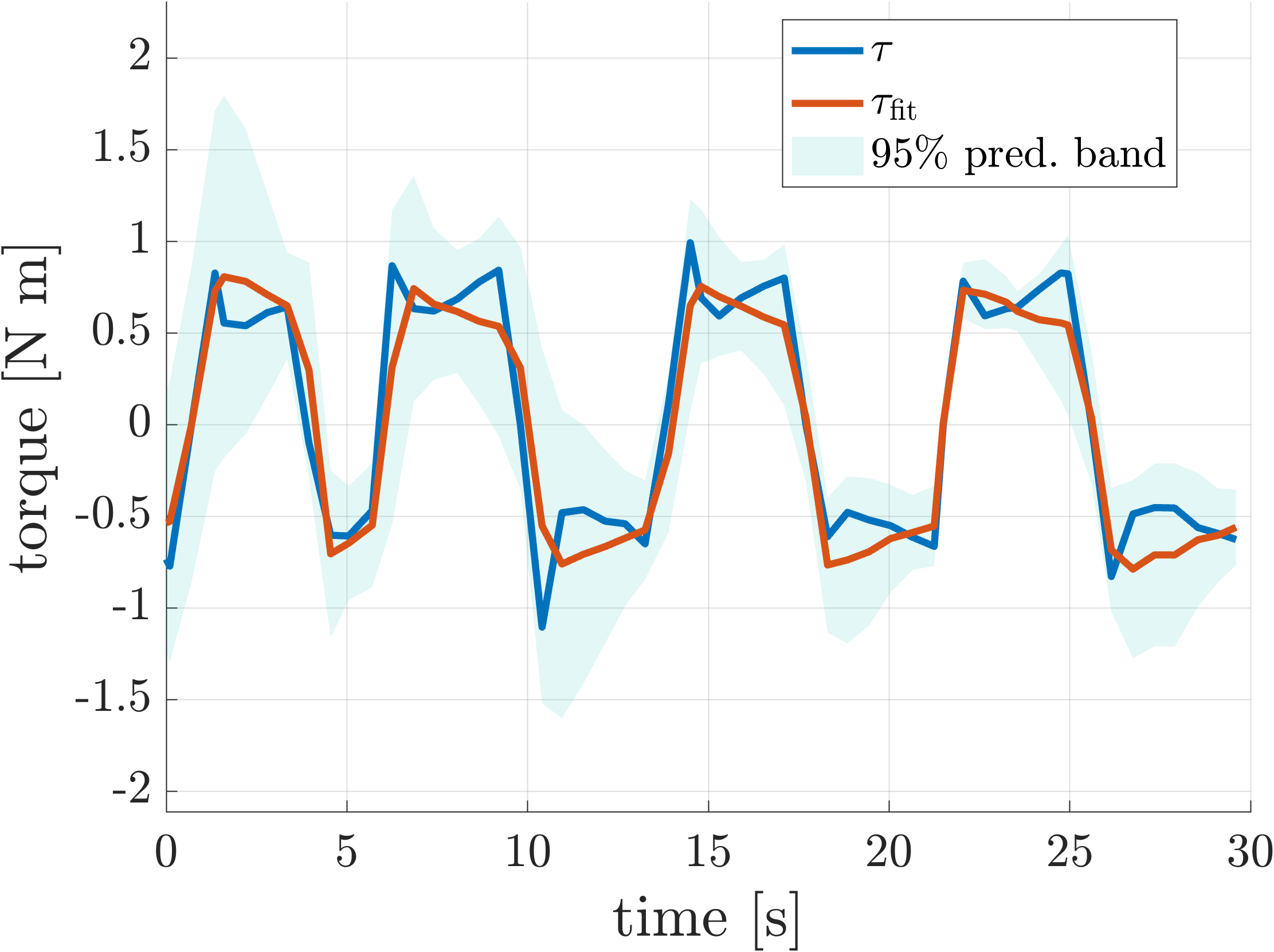}
  \end{subfigure}
  \begin{subfigure}{0.4\textwidth}
    \centering   \includegraphics[width=\linewidth]{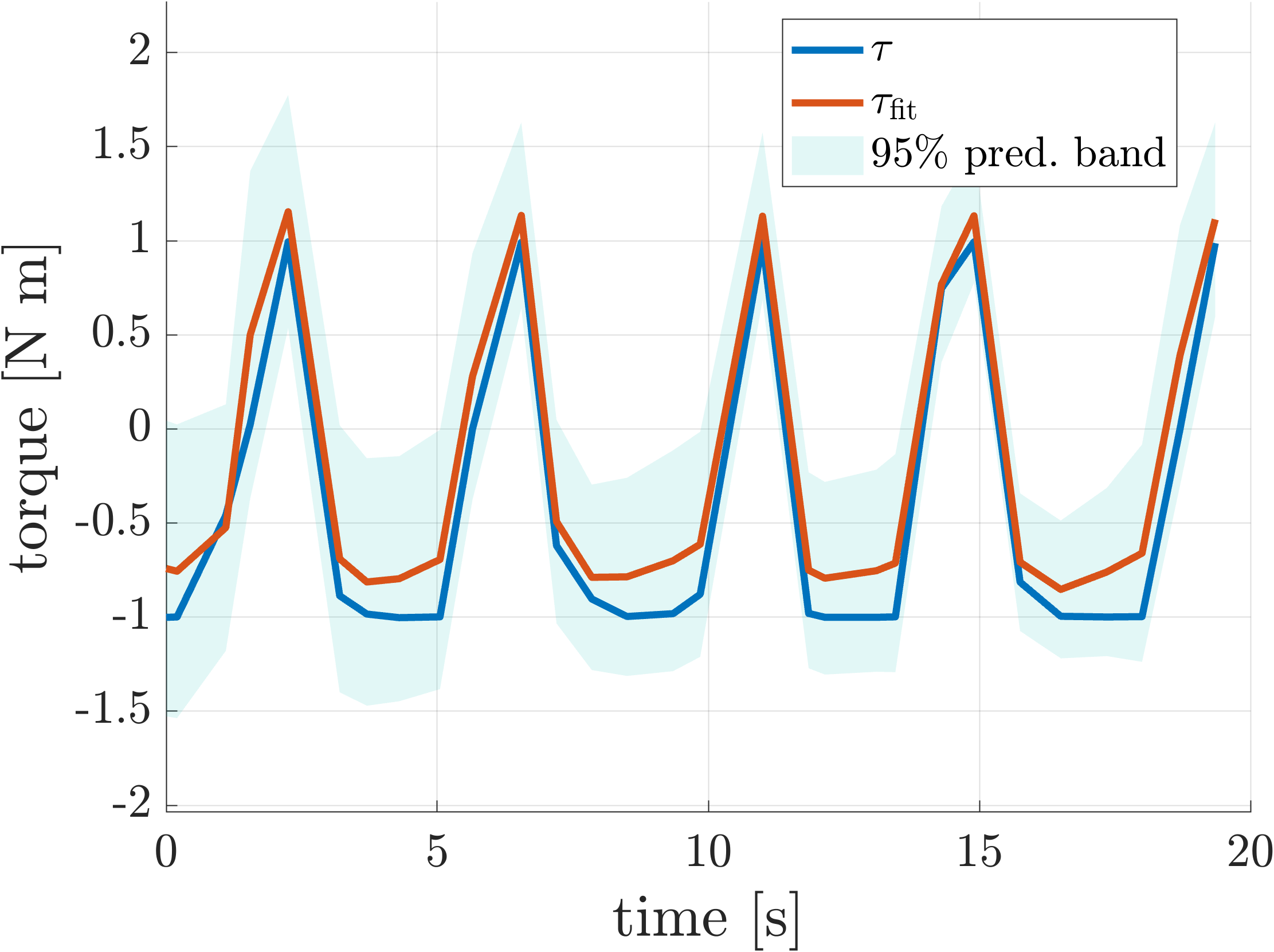}
  \end{subfigure}
  \caption{Identified fit for excited and recorded torque profile of manipulator: (a) joint 2  and  (b) joint 3.}
  \label{fig:fit_joint_3}
\end{figure*}

\begin{figure}[H]
  \centering
  \includegraphics[width=0.35\textwidth, angle=0]{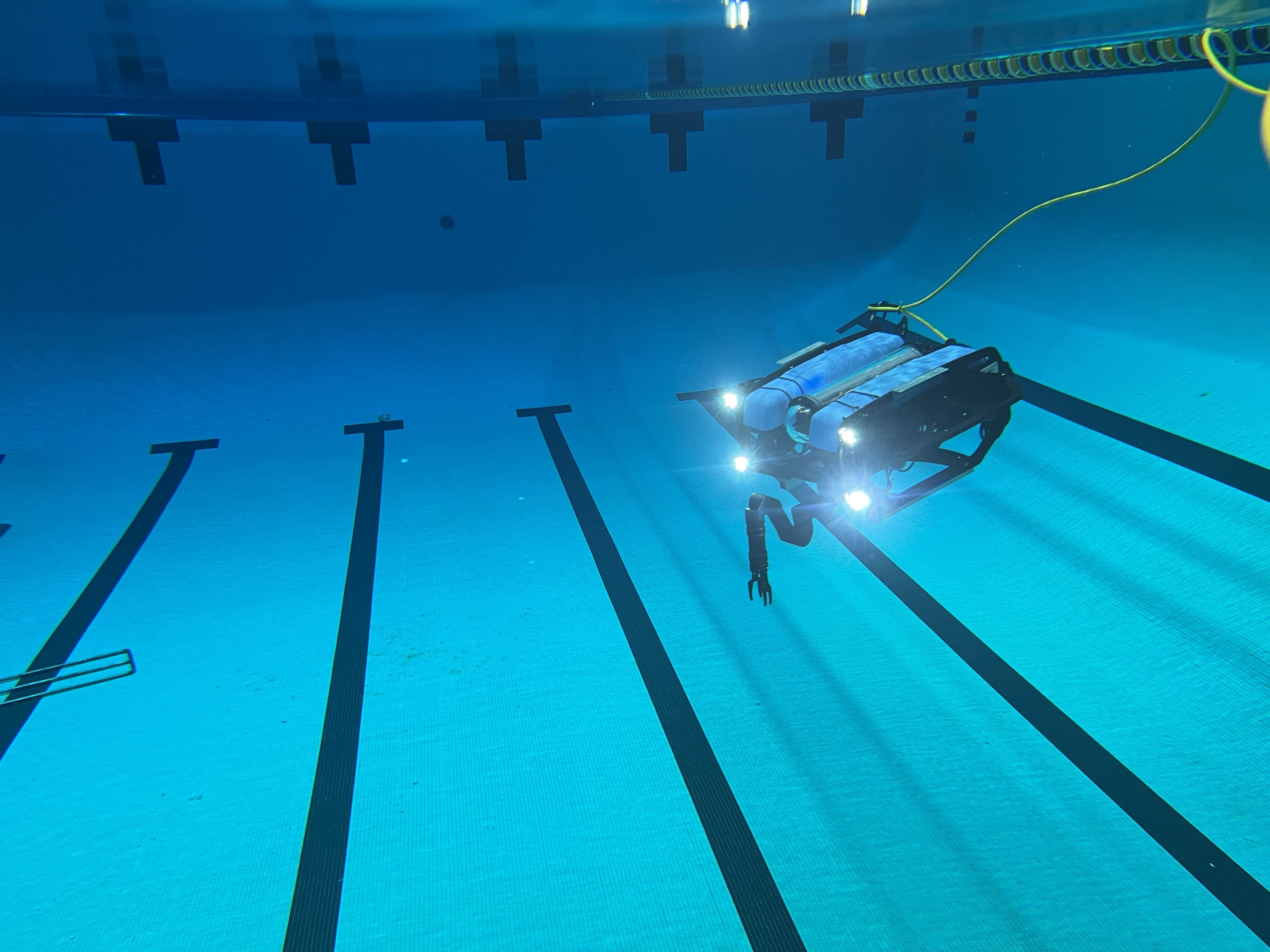}
  \caption{UVMS experiment platform.}
  \label{fig:platform_exp}
\end{figure}

Validation experiments were conducted in a 50~m$^2$ test pool using a 6-DOF BlueROV2 Heavy underwater vehicle coupled with a 4-DOF Reach Alpha 5 manipulator, as shown in Fig.~\ref{fig:platform_exp}. The vehicle was equipped with a depth/pressure transducer, a Doppler velocity log (DVL), and an inertial measurement unit (IMU). Sensors provided partial measurements of depth, orientation, body-fixed velocities, joint angles, and joint velocities. Acceleration signals were estimated in real time, filtered to attenuate measurement noise, and streamed to a station computer for use in online estimation and model updates. The station processor was an Intel\textsuperscript{\textregistered} Core\textsuperscript{TM} Ultra 9 275HX CPU with 64~GB DDR5 RAM, and the full estimation pipeline was implemented in ROS~2 Jazzy in Python and C++. The convex optimization problems, parameterized by a total of $k + 12n = 75$ unknowns ($k=27$ vehicle parameters, with $k_m=10$ inertial terms, $k_d=12$ damping terms, and $k_g=5$ restoring-force terms, together with $12n=48$ manipulator parameters for $n=4$ joints), were solved using MOSEK with warm start enabled, ensuring efficient solver performance during online adaptation.
The full implementation of the regressor construction, adaptive estimation framework, and the identified parameter sets used in these experiments are publicly available at 
\url{https://github.com/RKinDLab/floating-KinDyn}.

It is important to emphasize that the initial parameters of the model were deliberately set far from the true values of the physical system. The purpose of the validation experiment was therefore not to test a tuned model but to evaluate whether the estimation scheme could adapt parameters online and converge toward physically consistent values. Validation was carried out by comparing predicted torques from the identified model against the observed force and torque profiles, with progressive reduction in model–data error demonstrating the effectiveness of the adaptive update.



\begin{figure*}[htb]
\centering

\begin{subfigure}[t]{0.55\textwidth}
    \centering
    \includegraphics[width=\textwidth]{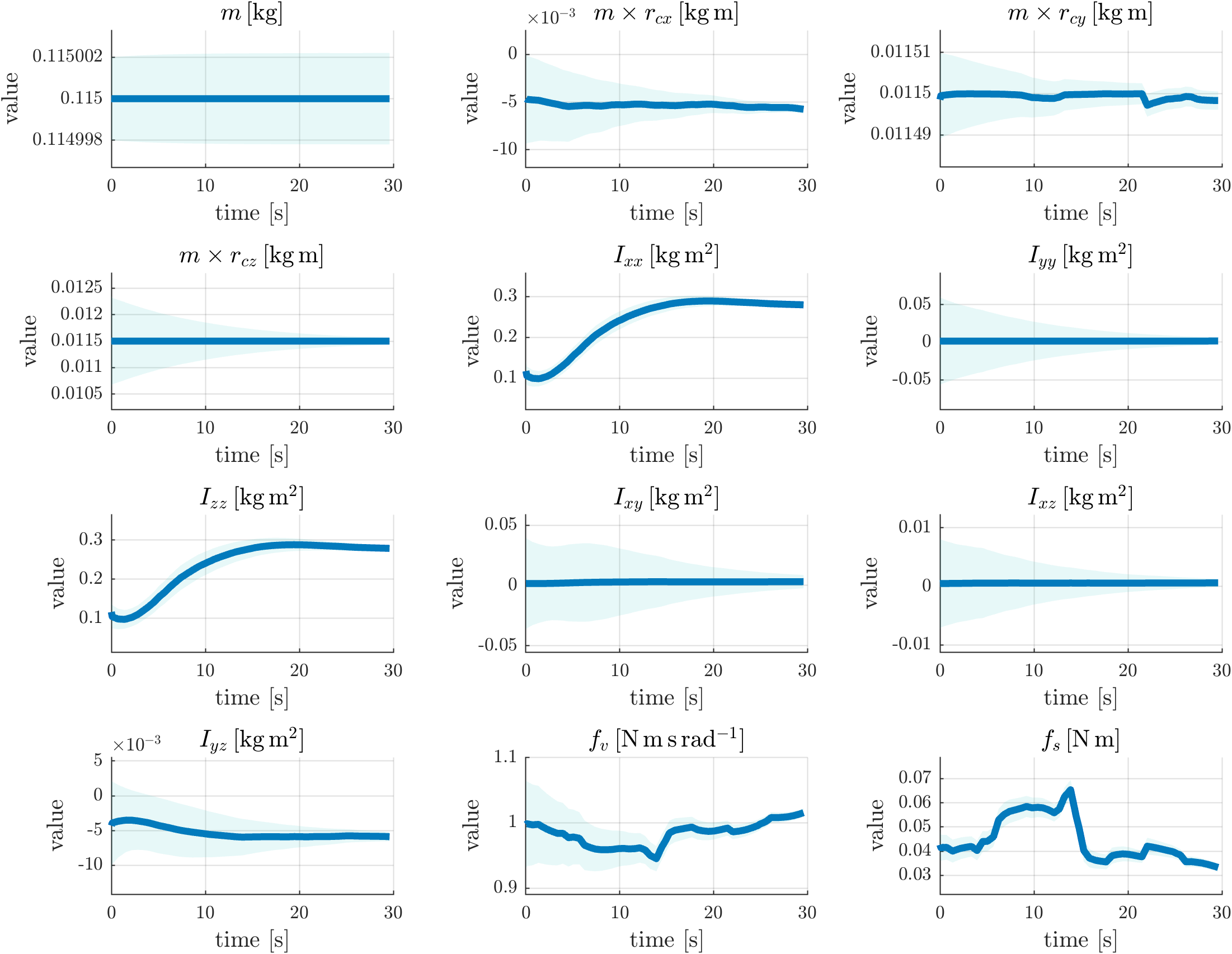}
    \caption{Joint 2 parameter evolution with 95\% confidence intervals.}
    \label{fig:manipulator_params_joint_2}
\end{subfigure}
\hfill
\begin{subfigure}[t]{0.42\textwidth}
    \centering
    \includegraphics[width=\textwidth]{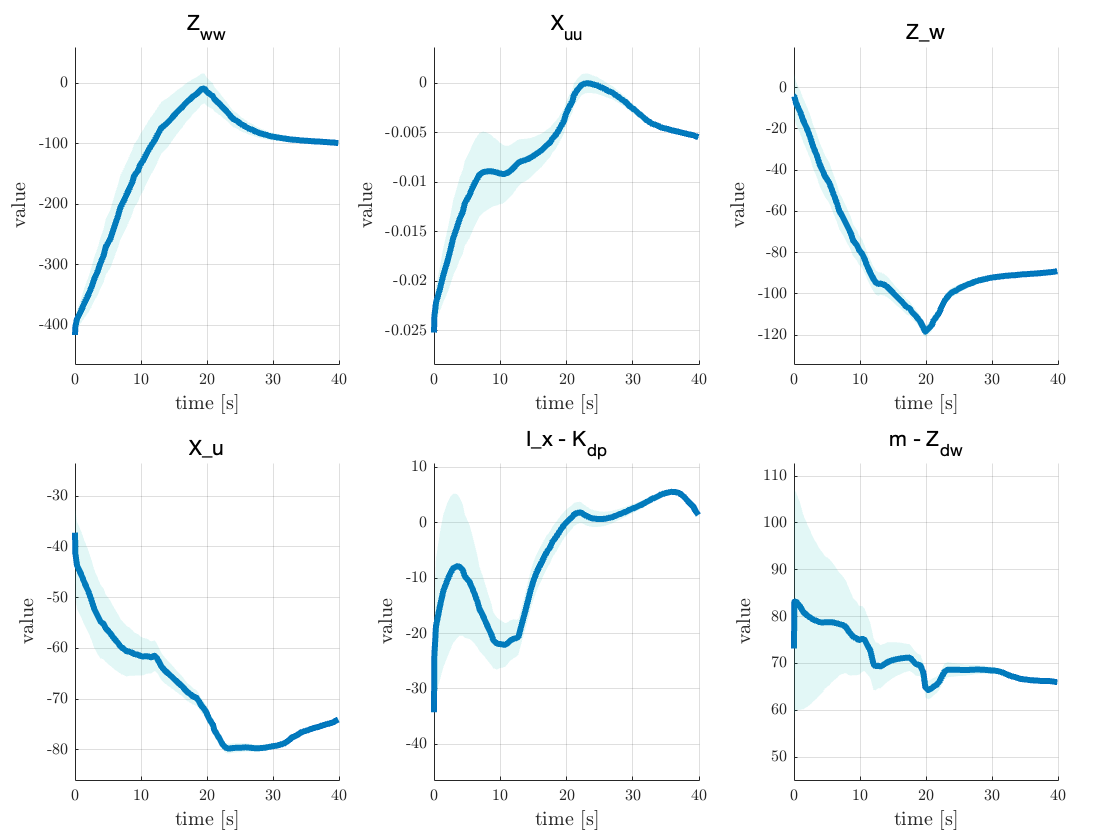}
    \caption{Selected vehicle parameters during online adaptation with 95\% confidence intervals.}
    \label{fig:selected_vehicle_params}
\end{subfigure}

\caption{Online parameter adaptation results for manipulator and vehicle subsystems.}
\label{fig:parameter_adaptation}
\end{figure*}

To sufficiently excite the coupled dynamics, vehicle thrusters and manipulator joints were constantly under excitation. Filtered outputs (vehicle and manipulator states) were logged synchronously and used to construct regressor matrices for the vehicle and manipulator. For the manipulator subsystem, the block upper-triangular structure of the regressor was explicitly exploited. The most distal joint, which depends only on its own parameters, was excited and solved first, and the solution then propagated proximally toward the base. For the vehicle subsystem, a staged excitation strategy was adopted to exploit the sparsity of the regressor. Yaw and heave \acp{DOF}, which exhibit minimal cross-coupling, were excited first to identify their respective inertial and damping parameters. Roll and pitch were then excited individually to isolate their diagonal dynamics, followed by combined surge–pitch and sway–roll maneuvers to reveal cross-inertia and coupling terms. Finally, quasi-static tilting experiments provided steady-state restoring moments, allowing estimation of the effective buoyancy distribution and center-of-gravity to center-of-buoyancy offsets. By freezing previously identified parameters and unlocking only small parameter subsets at each stage, the regressor remained well-conditioned and the resulting estimates remained physically consistent.  Together, this hierarchical identification procedure, sequentially distal-to-proximal for the manipulator and least-coupled-to-most-coupled for the vehicle, yielded a physically consistent set of parameters. Representative fits of joint torques and vehicle forces and moments are presented in the following to illustrate the accuracy of the identified models and the plausibility of the estimated parameters.


\begin{figure*}[ht]
  \centering
  \begin{subfigure}{0.4\textwidth}
    \centering  \includegraphics[width=\linewidth]{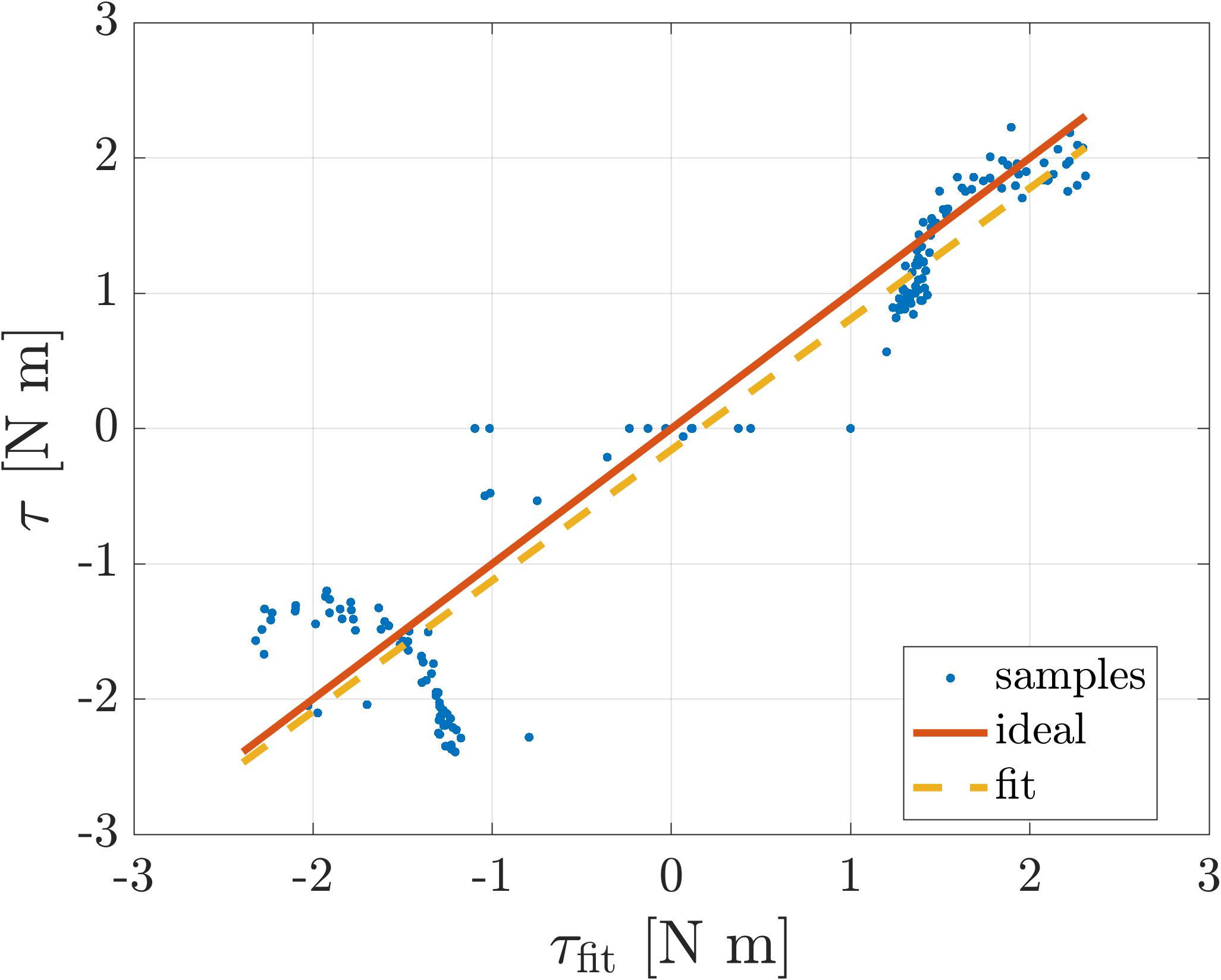}
    \caption{}
    \label{fig:params_joint_2_dup}
  \end{subfigure}
  ~
  \begin{subfigure}{0.4\textwidth}
    \centering
    \includegraphics[width=\linewidth]{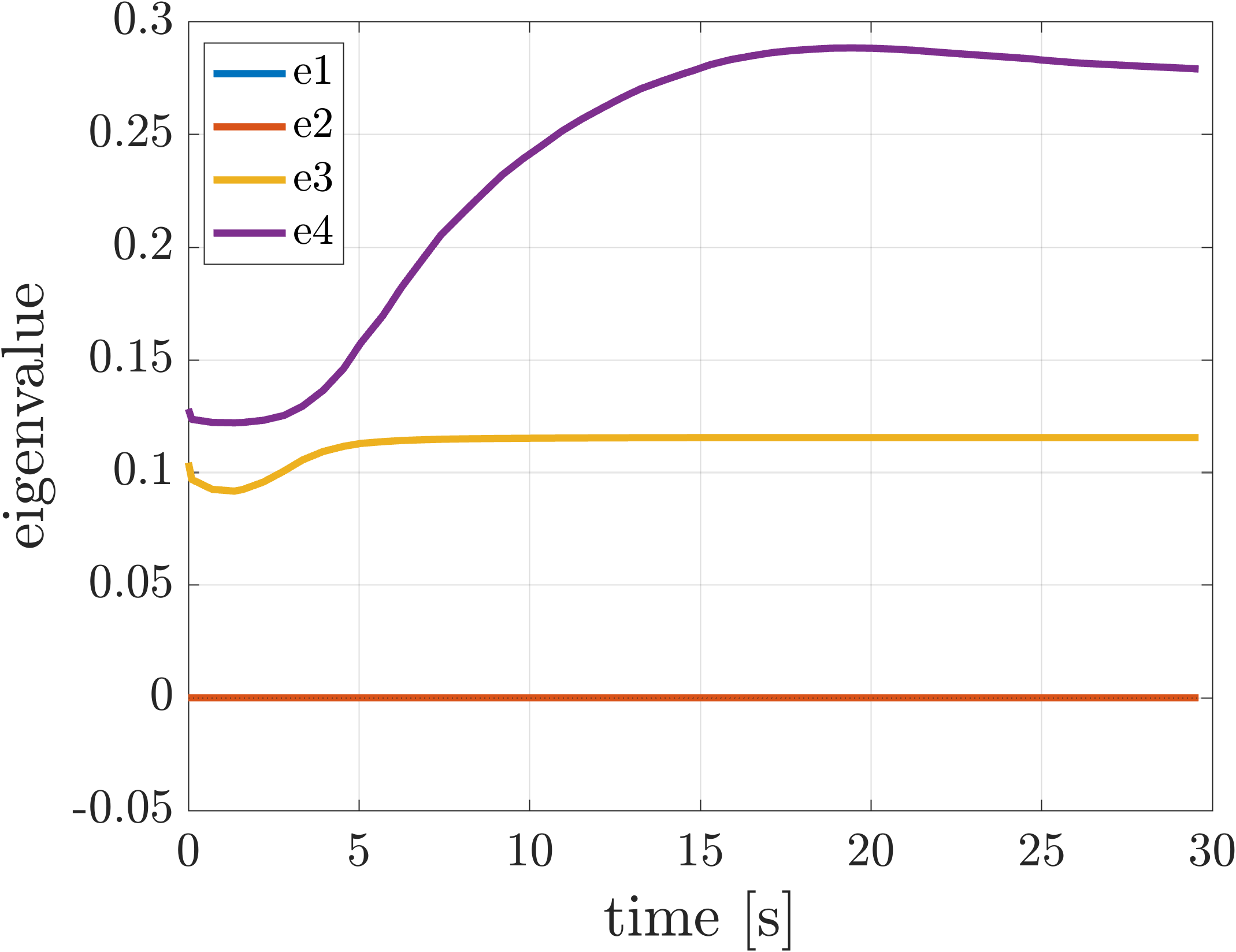}
    \caption{}
    \label{fig:eigs_joint_2}
  \end{subfigure}
  \caption{(a) Joint 0 parity analysis, excited torque versus model computed torque. (b) Eigenvalue trajectory of the pseudo inertia for joint 2, showing positive definiteness.}
\end{figure*}

\section{Results and Discussion}

The identified models achieve accurate torque, moment, and force prediction across both subsystems, while preserving physical consistency. The evaluation of the system has been done qualitatively and quantitatively. The metrics used are coefficient of determination ($R^2$), the \ac{MSE}, \ac{MAE},  and slope.

For the manipulator, torques shown in \figurename~\ref{fig:fit_joint_0} and \figurename~\ref{fig:fit_joint_3} closely follow the excited and recorded profiles ($\tau$), with slopes near unity and $R^2$ between $0.88$ and $0.98$. Joint~3 attains the best performance ($R^2=0.98$, slope $=1.01$, RMSE $\approx 0.22$~N$\cdot$m). Joints~0 to~2 also show strong agreement, with slopes $0.76$ to $0.97$ and RMSE below $0.5$~N$\cdot$m. The manipulator parameters converge rapidly, often within the first $10$s depending on excitation, as seen in \figurename~\ref{fig:manipulator_params_joint_2}. In this figure, it can be seen that the uncertainty bands contract after this initial adaptation converging to the true value of the parameters. Furthermore, by studying the eigenvalues of the pseudo-inertia matrix of the manipulator links (\figurename~\ref{fig:eigs_joint_2}), the matrix is confirmed to be positive definite and the validity of the inertia parameters can be verified.

Identifying the vehicle parameters is  more challenging due to hydrodynamic variability and stronger cross coupling effects between \acp{DOF}. Behaviors in surge, heave, and roll are reproduced with moderate fidelity, as seen in  \figurename~\ref{fig:vehicle-prediction}, while sway and pitch show lower accuracy. Roll dynamics are captured with $R^2=0.72$ and slope $=1.13$. The vehicle requires more time and richer excitation across its DOFs to reach stable convergence, and higher sensitivity to sensor noise appears in wider confidence intervals and small oscillations that narrow with repeated excitation, as shown in \figurename~\ref{fig:selected_vehicle_params}.

The lower fidelity in sway and pitch is mainly attributed to comparatively weaker and less independent excitation of these DOFs during the identification experiments. Although dedicated excitation sequences were applied, surge, heave, and yaw motions were easier to isolate and excite with higher amplitude and broader frequency content, resulting in stronger persistence of excitation in the corresponding regressor blocks. In contrast, sway motion remained more strongly coupled to roll and yaw, and the lateral hydrodynamic forces were influenced by thruster-induced flow interactions and pool boundary effects. These factors reduce the effective signal-to-noise ratio and make isolation of pure sway dynamics more difficult.

Pitch dynamics present a similar challenge. While pitch was intentionally excited, its behavior is strongly dominated by restoring moments arising from buoyancy distribution and center-of-gravity to center-of-buoyancy offsets. Small uncertainties in these geometric parameters can produce noticeable discrepancies in predicted moments. Moreover, pitch responses are typically lower in amplitude compared to surge or heave under safe experimental constraints, which limits independent excitation of inertial and damping terms. Consequently, parameter observability in sway and pitch remains weaker relative to the more directly and persistently excited DOFs, leading to reduced predictive fidelity despite physically consistent convergence.

\begin{table}[H]
\centering
\caption{Average identification metrics for vehicle DOFs and manipulator joints, based on recorded excited profiles. }
\label{tab:metrics}
\begin{tabular}{lccccc}
\toprule
DOF / Joint & $R^2$ & slope & MSE & MAE & RMSE \\
\midrule
surge (N)   & 0.58 & 0.62 & 16.081 & 2.939 & 4.010\\
sway (N)    & 0.46 & 1.08 & 20.072 & 3.378 & 4.480 \\
heave (N)   & 0.68 & 1.02 & 42.658 & 3.397 & 6.531 \\
roll (N\,m) & 0.72 & 1.13 & 1.085 & 0.730 & 1.041 \\
pitch (N\,m)& 0.43 & 0.51 & 1.206 & 0.879 & 1.098 \\
yaw (N\,m)  & 0.68 & 1.06 & 0.392 & 0.439 & 0.626 \\
joint 0 (N\,m) & 0.90 & 0.96 & 0.264 & 0.525 & 0.007 \\
joint 1 (N\,m) & 0.88 & 0.76 & 0.136 & 0.026 & 0.477 \\
joint 2 (N\,m) & 0.89 & 0.97 & 0.160 & 0.348 & 0.004 \\
joint 3 (N\,m) & 0.98 & 1.01 & 0.077 & 0.071 & 0.220 \\
\bottomrule
\end{tabular}
\end{table}

Error metrics corroborate these trends, as seen in Table~\ref{tab:metrics}. 
Results are averaged over 10 experimental trials, each lasting approximately 40~s. 
In these experiments, the 95\% predictive intervals consistently enveloped all recorded torque and force profiles across all runs. For the manipulator, MSE remains below $0.15~(\mathrm{N\cdot m})^2$, MAE ranges from $0.05$ to $0.1~\mathrm{N\cdot m}$, and RMSE is at or below $0.6~\mathrm{N\cdot m}$. 
As expected, $\mathrm{RMSE} \ge \mathrm{MAE}$ due to the quadratic penalization of larger residuals. For the vehicle, larger discrepancies and broader uncertainty intervals align with higher noise sensitivity and stronger cross-coupling effects.


\begin{figure}[ht]
\centering  \includegraphics[width=0.4\textwidth]{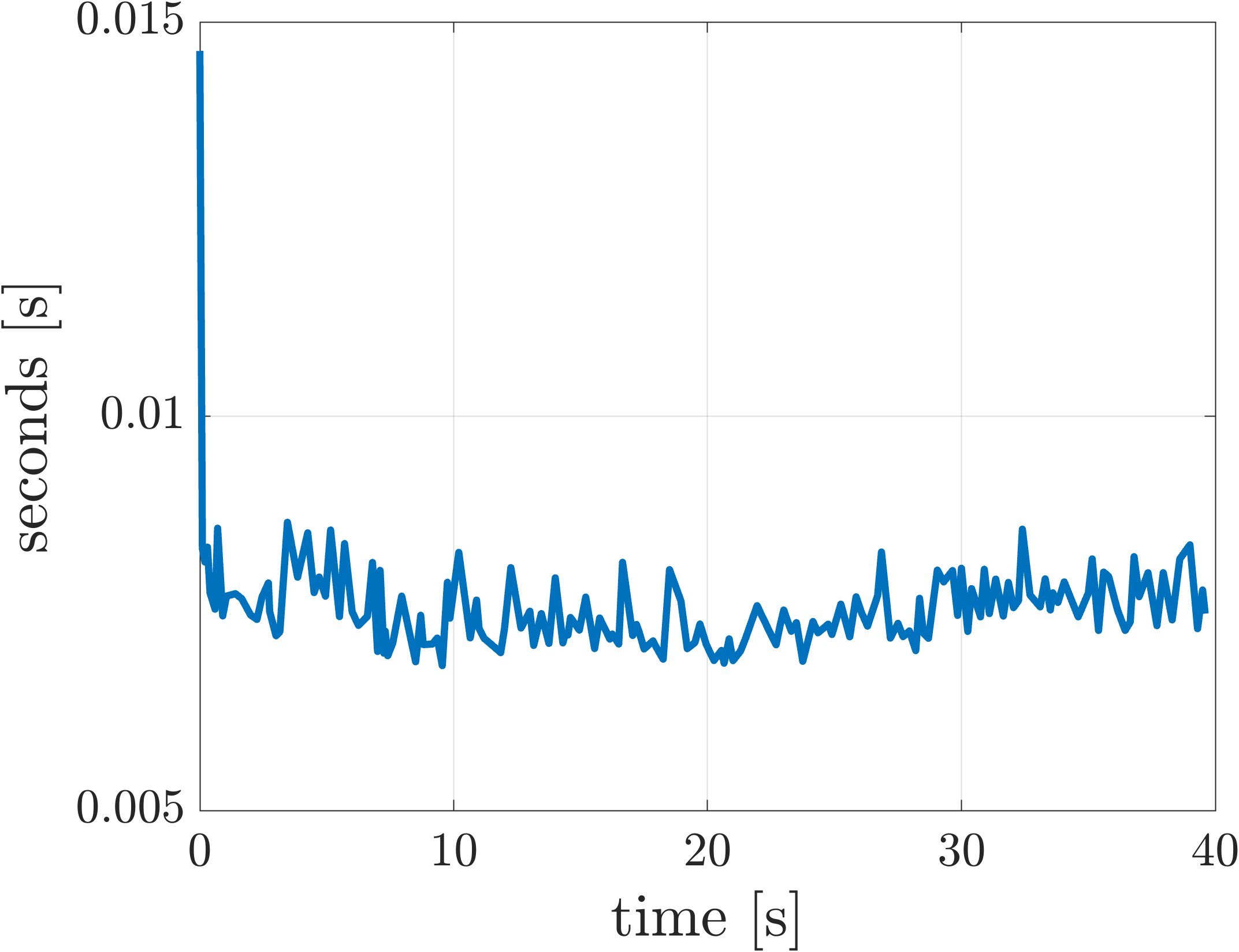}
  \caption{Solver time during online identification.}
  \label{fig:solver-time}
\end{figure}

For these experiments, solver times remain below $0.03$~s per update, with a median of $\approx 0.023$~s, seen in \figurename~\ref{fig:solver-time}. This confirms the computational feasibility of imposing convex physical consistency constraints during online adaptation.

\begin{figure}[ht]
  \centering
  \includegraphics[width=0.495\textwidth]{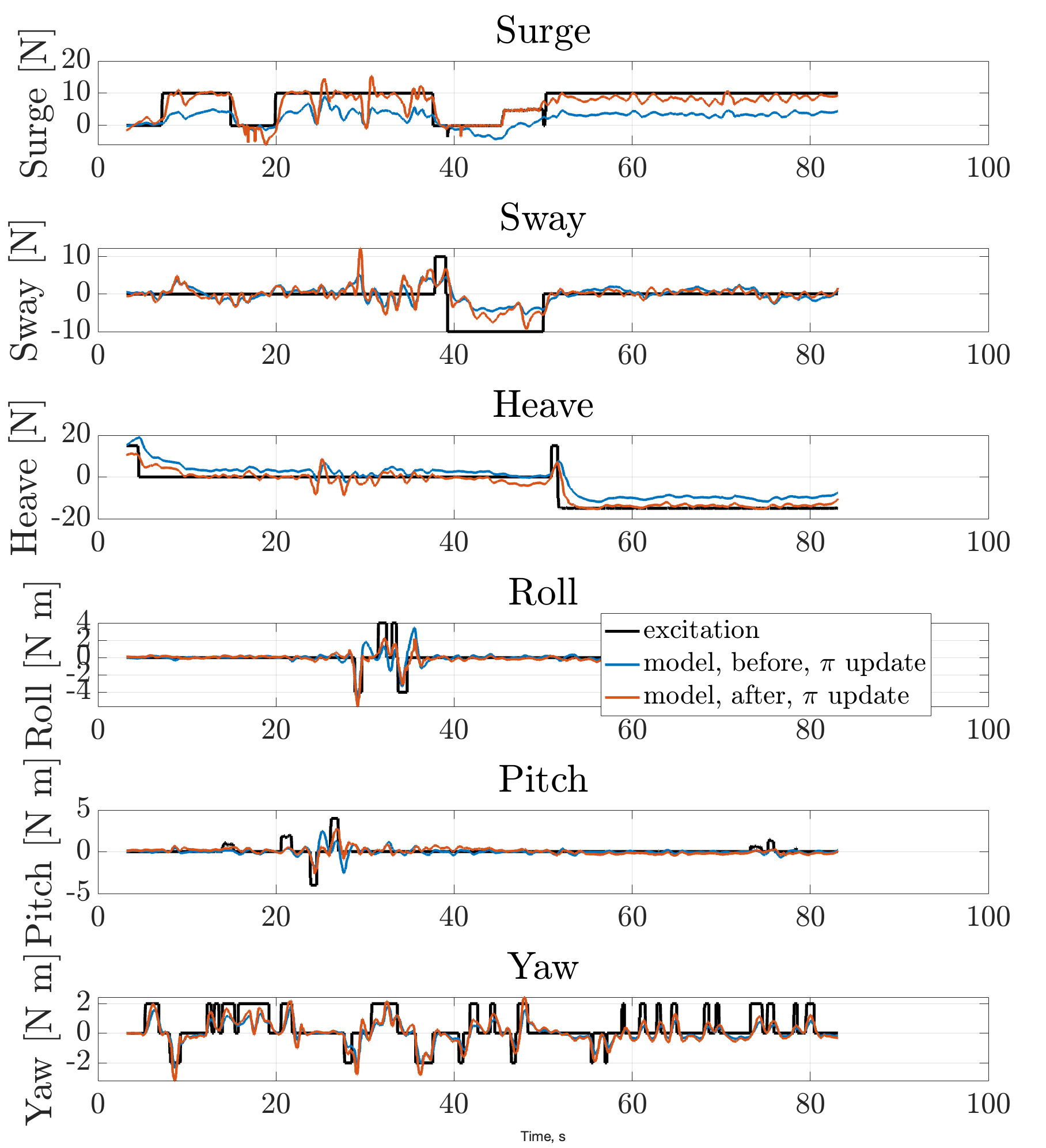}
  \caption{Comparison of excitation torque versus model torques computed with prior fixed parameters (blue plots)  and with the adaptive converged parameters (red plots), for vehicle DOFs. 
  }
  \label{fig:compare_fixed}
\end{figure}
To further quantify the benefit of the proposed adaptive parameter estimation, the retrieved inverse dynamic model was evaluated offline against a prior model, and the resulting computed torques were compared against the excitation forces, moments and torques on each of the robot DOF. The mean comparison in Fig.~\ref{fig:compare_fixed} shows consistent improvement across DOFs, lower MAE, lower RMSE, and smaller signed error, which aligns with the gains reported in Table~\ref{tab:metrics}. A directional asymmetry is visible on the reference system DOFs, where trajectories differ slightly for positive versus negative torque, consistent with hysteresis-like behavior or direction-dependent friction. Likely contributors include asymmetric Coulomb plus Stribeck friction in joints, gearbox backlash, thruster deadband, and thrust curve asymmetry under flow reversal. Incorporating asymmetric friction terms, a small deadband, or a simple hysteresis element, or splitting identification by torque sign, could further reduce these residuals while preserving physical consistency. Overall, the identified models yield physically plausible parameters and calibrated, uncertainty-aware predictions, making them suitable for real-time control and planning in underwater manipulation.

\vspace{-7pt}
\section{Conclusion}
\label{sec:conclusions}
The paper presented an approach to adaptively estimate the parameters of \acfp{UVMS}. The proposed approach can learn the parameters and their uncertainty online, based on the states and inputs of the system and on a moving horizon estimation formulation.  The paper introduces, for the first time, a unified regressor method for  the \ac{UVMS} dynamics.  Such an approach is useful for the creation of accurate dynamic digital-twin environments, the development of control systems, and improving the models used for localization of underwater vehicles. Future work will focus on the implementation of such a model for energy-efficient model-based control architectures.

\bibliography{library} 
\bibliographystyle{IEEEtran}

\end{document}